\newcommand{\CLASSINPUTtoptextmargin}{1.92cm}
\newcommand{\CLASSINPUTbottomtextmargin}{2.53cm}

\documentclass[conference]{IEEEtran}
\IEEEoverridecommandlockouts
\usepackage{tabu}
\usepackage{extarrows}
\usepackage{xspace}
\usepackage{booktabs}
\usepackage{multirow}
\usepackage{array}
\usepackage{graphicx}
\usepackage{wrapfig}
\usepackage{float}
\usepackage{booktabs}
\usepackage{threeparttable}
\usepackage{pifont}
\newcommand{\cmark}{\ding{51}}

\usepackage{url}
\usepackage{hyperref}

\usepackage{textcomp}

\usepackage{picins/picins}

\usepackage{cite}
\usepackage{amsmath,amssymb,amsfonts}
\usepackage{algorithmic}
\usepackage{graphicx}
\usepackage{textcomp}
\usepackage{soul}
\usepackage[dvipsnames]{xcolor}
\def\BibTeX{{\rm B\kern-.05em{\sc i\kern-.025em b}\kern-.08em
    T\kern-.1667em\lower.7ex\hbox{E}\kern-.125emX}}

\newif\ifcomment

\commenttrue

\ifcomment
\newcommand{\giannis}[1]{\sethlcolor{pink}\hl{[\textbf{Giannis:} #1]}}
\newcommand{\sokratis}[1]{\sethlcolor{SkyBlue}\hl{[Sokratis: #1]}}

\else
\newcommand{\giannis}[1]{}
\newcommand{\sokratis}[1]{}

\fi

\makeatletter
\def\footnoterule{\relax%
  \kern-5pt
  \hbox to \columnwidth{\hfill\vrule width 1\columnwidth height 0.4pt\hfill}
  \kern4.6pt}
\makeatother

\begin{document}

\title{Exploring the Performance and Efficiency of Transformer Models for NLP on Mobile Devices}

\author{\IEEEauthorblockN{
Ioannis Panopoulos\IEEEauthorrefmark{2},
Sokratis Nikolaidis\IEEEauthorrefmark{2},
Stylianos I. Venieris\IEEEauthorrefmark{3},
Iakovos S. Venieris\IEEEauthorrefmark{2}
}
\IEEEauthorblockA{
\IEEEauthorrefmark{2}National Technical University of Athens, Athens, Greece,
\IEEEauthorrefmark{3}Samsung AI Center, Cambridge, UK}
\IEEEauthorblockA{Email: ioannispanop@mail.ntua.gr, sokratisnikolaidis@mail.ntua.gr, s.venieris@samsung.com, venieris@cs.ece.ntua.gr}
\vspace{-0.8cm}
}

\maketitle

\begin{abstract}
    Deep learning (DL) is characterised by its dynamic nature, with new deep neural network (DNN) architectures and approaches emerging every few years, driving the field's advancement. At the same time, the ever-increasing use of mobile devices (MDs) has resulted in a surge of DNN-based mobile applications. Although traditional architectures, like CNNs and RNNs, have been successfully integrated into MDs, this is not the case for Transformers, a relatively new model family that has achieved new levels of accuracy across AI tasks, but poses significant computational challenges. In this work, we aim to make steps towards bridging this gap by examining the current state of Transformers' on-device execution. To this end, we construct a benchmark of representative models and thoroughly evaluate their performance across MDs with different computational capabilities. Our experimental results show that Transformers are not accelerator-friendly and indicate the need for software and hardware optimisations to achieve efficient deployment.
\end{abstract}


\vspace{-0.15cm}
\section{Introduction}
\label{sec:intro}
Being the cornerstone of many artificial intelligence (AI) tasks, DNNs are rapidly making their way into user-facing mobile applications~\cite{dlsmartphones2019www}, imposing an increasing demand for real-time inference with low latency and high accuracy. To achieve this, different execution options have been proposed, including running DNN inference in the cloud, at the edge, and on-device~\cite{dledge2019pieee}. While cloud execution offers high performance, it requires a stable network connection, and edge execution involves higher latency due to data transfer. On-device execution, on the other hand, enables real-time inference without network requirements, making it an attractive option for applications that require low latency, privacy, and security.

For the past several years, convolutional neural networks (CNNs) were the dominant DNN architecture and, thus,
have been extensively researched and optimised (\textit{e.g.}~with techniques such as the depthwise separable convolution) for various applications, leading to efficient on-device inference~\cite{efficientnet2019icml}.
Since the introduction of BERT~\cite{bert2019naacl}, however, Transformers have gained immense popularity not only in natural language processing (NLP) tasks~\cite{t52020jmlr, roberta2019arxiv}, 
but also in domains such as computer vision~\cite{vit2021iclr} and speech recognition~\cite{transformers_speech202interspeech}. With such versatility, there is an upcoming requirement for Transformer-driven applications and their deployment close to the user. Even so, the current focus of the research community is primarily on server-based training and inference~\cite{mobius2023asplos}, leaving the on-device execution of Transformers largely unexplored.



In this work, we argue that well-established practices and findings for the on-device execution of CNNs cannot translate directly to Transformer models.
This paper makes the following key contributions:
\begin{itemize}
    \item A benchmark suite of diverse Transformer models and the associated software infrastructure, which enables on-device evaluation in a systematic and reproducible way.
    \item A thorough investigation of the current state of on-device Transformer inference, which includes benchmarking a wide range of models, exploring the compatibility of various mobile processors, validating on-device accuracy, and analysing the suitability of various quantisation methods.
    \item Suggestions for future optimisations,
    which can increase hardware compatibility and reduce computational needs.
\end{itemize}

\section{Background}
\label{sec:background}
\subsection{Challenges of On-Device DNN Inference}
\label{sec:background_ondev_exec}
Despite the benefits of on-device execution, there are several challenges that need to be addressed. MDs have limited resources, while at the same time, on-device execution requires a significant amount of memory and processing power, due to the increasing complexity of neural network models. Moreover, the heterogeneity of MDs in terms of both software and hardware often requires the deployment of a given model to be manually tuned for each target device~\cite{smartcost2021imc}.


\begin{table*}[t!]
    \centering
    \caption{\small Transformer Models}
    \vspace{-0.2cm}
    \setlength{\tabcolsep}{4pt}
        \begin{tabu}{c | c r r | l r r r r r | r r | c c c c}
            \toprule
            

            \multirow{2}{*}{\textbf{Task}} & \multicolumn{3}{c|}{\textbf{Embeddings}} & \multicolumn{6}{c|}{\textbf{Encoder}} & \multicolumn{2}{c|}{\textbf{Overall Model}} & \multicolumn{4}{c}{\textbf{Accuracy (\%)}} \\
            
            & \textbf{Vocab} & \textbf{E} & \textbf{\#Params} & \textbf{Name} & \textbf{L} & \textbf{A} & \textbf{H} & \textbf{I} & \textbf{\#Params} & \textbf{FLOPs} & \textbf{\#Params} & \textbf{FP32} & \textbf{FP16} & \textbf{DR8} & \textbf{FX8} \\
            
            \midrule
                                             & 50265 & 24  & 1.21 M    & BERT Tiny        & 6  & 2  & 24  & 16   & 0.02 M   & 3.53 M  & 1.23 M   & 90.05 & 90.05 & 90.05 & 89.85 \\
                                             & 30522 & 128 & 3.91 M    & ELECTRA Tiny     & 6  & 2  & 24  & 16   & 0.03 M   & 4.19 M  & 3.94 M   & 90.25 & 90.30 & 90.15 & 90.50 \\
                                             & 30522 & 256 & 7.81 M    &XDistil-L6-H256& 6  & 8  & 256 & 1024 & 4.75 M   & 0.49 G  & 12.57 M  & 93.20 & 93.25 & 93.00 & 30.95 \\
            \textbf{Sequence}                & 30522 & 384 & 11.72 M   & MiniLM-L3     & 3  & 12 & 384 & 1536 & 5.35 M   & 0.54 G  & 17.07 M  & 93.25 & 93.25 & 93.00 & 91.95 \\
            \textbf{Classification}          & 30522 & 128 & 3.91 M    & MobileBERT    & 24 & 4  & 128 & 512  & 20.42 M  & 2.06 G  & 24.33 M  & 93.30 & 93.30 & 93.10 & - \\
            on Emotions                      & 30522 & 384 & 11.72 M   &XDistil-L6-H384& 6  & 12 & 384 & 1536 & 10.67 M  & 1.09 G  & 22.39 M  & 93.35 & 93.35 & 93.10 & 82.30 \\
            Dataset                      & 30522 & 384 & 11.72 M   & MiniLM-L12        & 12 & 12 & 384 & 1536 & 21.32 M  & 2.18 G  & 33.04 M  & 93.45 & 93.45 & 92.65 & 78.40 \\
                                             & 28996 & 512 & 14.85 M   & RoBERTa Tiny     & 4  & 8  & 512 & 2048 & 12.64 M  & 1.28 G  & 27.49 M  & 93.50 & 93.50 & 93.40 & 92.30 \\
                                             & 30522 & 128 & 3.91 M    & ELECTRA Small     & 12 & 4  & 256 & 1024 & 9.55 M   & 0.98 G  & 13.46 M  & 93.55 & 93.55 & 93.30 & 92.25 \\
                                             & 30522 & 768 & 23.44 M   & DistilBERT    & 6  & 12 & 768 & 3072 & 43.16 M  & 4.30 G  & 66.60 M  & 94.50 & 94.50 & 94.55 & 77.85 \\
            \midrule

            \textbf{Text}         & 50257 & 768 & 38.60 M   & DistilGPT2  & 6  & 12 & 768 & 3072 & 42.69 M  & 10.47 G & 81.29 M  & 46.85 & - & - & - \\
            \textbf{Generation}   & 50257 & 768 & 38.60 M   & GPT2 Small  & 12 & 12 & 768 & 3072 & 85.28 M  & 15.99 G & 123.88 M & 51.41 & 51.41 & - & - \\

            
            \bottomrule
        \end{tabu}
    \label{tab:transformers}
    \vspace{-0.45cm}
\end{table*}

\subsection{Transformers}
The architecture of a Transformer is typically composed of an encoder and a decoder~\cite{attention2017nips}. The encoder produces a 
vector representation of the input sequence, while the decoder generates the output sequence one token at a time, using the encoder's representation as context. In a typical flow, a Transformer's sequence-to-sequence architecture is first 
pre-trained on language modelling or next sentence prediction tasks and, then, the encoder can be fine-tuned to a wide range of downstream tasks~\cite{bert2019naacl}.
The architectural parameters that affect a Transformer model's latency during the inference stage are listed in Table~\ref{tab:arch_params}.

\begin{table}[h]
  \centering
  \vspace{-0.2cm}
  \caption{Transformer Architectural Parameters}
  \label{tab:arch_params}
  \vspace{-0.2cm}
  \resizebox{\columnwidth}{!}{
  \begin{tabular}{l | l}
    \toprule
    \textbf{Parameter} & \textbf{Description} \\
    \midrule
    Sequence Length (S) & Maximum number of tokens in one input sample. \\
    Embedding Size (E) & Width of a token’s embedding. \\
    Hidden Size (H) & Dimensionality of the model's internal vector space. \\
    Attention Heads (A) & Number of parallel heads for the attention mechanism. \\
    Intermediate Size (I) & Width of the FFN Network. \\
    Layers (L) & Number of layers in the model. \\
    \bottomrule
\end{tabular}
}
\vspace{-0.55cm}
\end{table}

\subsection{Mobile Processors}
\label{sec:background_mobile_processors}
Mobile processors are a core component of modern smartphones, tablets and other portable devices,
as they directly impact factors such as battery life and response time.
In the context of DNN inference, the central processing unit (CPU) is the baseline processor, while the rest are considered accelerators, as they can provide execution speedup and energy efficiency. The most common mobile accelerators at the moment are the graphics processing unit (GPU), the digital signal processor (DSP) and  neural processing units (NPUs)~\cite{accelerators2021cdt}.


\vspace{-0.15cm}
\section{Related Work}
\label{sec:related}
\vspace{-0.15cm}
On-device execution and evaluation of Transformer models for NLP remains largely unexplored. Previous works~\cite{reddi2020mlperf,mobile_accel2021emdl} have limitations, as they consider a small number of models, lack on-device accuracy evaluation and compression applicability, and do not consider all possible accelerators and execution configurations. Conversely, extensive research has been conducted to benchmark Vision Transformers and compare them to CNNs for use on MDs~\cite{ZAIDI2022103514, WaZhLiYa2022}, thus we do not include them in our work.

\vspace{-0.15cm}
\section{Experimental Methodology}
\label{sec:setup}
\vspace{-0.1cm}
For our evaluation infrastructure, we chose \mbox{TensorFlow} (\textit{v2.11.0}), because of its extensive range of quantisation methods and its support for mobile deployment through TensorFlow Lite (\textit{nightly} builds). For the benchmarked models, we used Hugging Face's Hub and Transformers~(\textit{v4.27.4}). Our experiments were conducted using Android devices and a custom-built Android application with a simple UI in order to examine how a model would perform in a real-life scenario.

\vspace{-0.05cm}
\subsection{Models and Tasks}
\vspace{-0.05cm}
Table~\ref{tab:transformers} presents the benchmarked Transformer models along with their architectural parameters, workload, size, and accuracy. 
In order to further optimise our models, we applied three of TFLite's post-training quantisation\footnote{\href{https://www.tensorflow.org/lite/performance/post_training_quantization}{https://www.tensorflow.org/lite/performance/post\_training\_quantization}} methods: half-precision floating-point (FP16), 8-bit dynamic range (DR8), and 8-bit fixed-point (FX8). Quantisation is one of the simplest and fastest compression methods presently available, with benefits not only in model size, but also latency and memory.

For our downstream classification task, we used Emotions, a small dataset consisting of English Twitter messages with six emotions as classes.
The reported accuracy corresponds to the top-1 accuracy on the dataset's test set of 2000 samples when using 50 tokens as the input sequence length.
We obtained the majority of models pre-trained on various large datasets, such as Reddit comments and 2ORC citation pairs, and subsequently fine-tuned them on Emotions. Selected models include optimised architectures, such as XtremeDistil~\cite{xtremedistiltransformers2021arxiv}, MiniLM~\cite{minilm2020nips} and MobileBERT~\cite{mobilebert2020acl}, as well as light versions of the RoBERTa~\cite{roberta2019arxiv} and ELECTRA~\cite{electra2020iclr} original implementations. Exceptions to these are DistilBERT~\cite{distilbert2019emc2}, whose Emotions-fine-tuned version was obtained directly from Hugging Face, and BERT and ELECTRA Tiny which were trained from scratch. The training/fine-tuning configuration involves a batch size of 64 and the RMSProp optimiser with initial learning rate in the order of $10^{-4}$ and exponential decay.


In addition to the above models, we also include GPT2~\cite{gpt22019} for text generation in its small and distilled versions, which we obtained already converted into the \texttt{tflite} format from Hugging Face.
We report accuracy in the context of the next token prediction task using LAMBADA's test set of 5000 passages and 64 input sequence and output prediction tokens.

\subsection{Mobile Devices}
To cover devices with different capabilities, we target two smartphones: Samsung Galaxy A71, representing the mid-tier category, and Samsung Galaxy S20 FE, representing the high-end category of modern mobile phones (Table~\ref{tab:devices}).

\begin{table}
  \centering
  \vspace{-0.2cm}
  \caption{Target Smartphones}
  \label{tab:devices}
  \vspace{-0.2cm}
  \resizebox{\columnwidth}{!}{
  \begin{tabular}{r | l | l}
    \toprule
    \textbf{Device} & 
    \textbf{Samsung A71} & 
    \textbf{Samsung S20 FE} \\
    \midrule
    \textbf{Date} & 2020, January & 2020, October \\
    \textbf{SoC} & Snapdragon 730 & Exynos 990 \\
    \multirow{3}{*}{\textbf{CPU}} & \multirow{2}{*}{2$\times$2.2 GHz Kryo 470 Gold} & 2$\times$2.73 GHz Exynos M5 \\
    & \multirow{2}{*}{6$\times$1.8 GHz Kryo 470 Silver} & 2$\times$2.5 GHz Cortex-A76 \\
    & & 4$\times$2.0 GHz Cortex-A55 \\
    \textbf{GPU} & Adreno 618 @700 MHz & Mali-G77 MP11 @800 MHz \\
    \textbf{NPU} & Qualcomm Hexagon 688 & \cmark \\
    \textbf{RAM} & 6 GB @1866 MHz & 6 GB @2750 MHz \\
    \textbf{TDP} & 5 W & 9 W \\
    \bottomrule
\end{tabular}
}
\vspace{-0.55cm}
\end{table}

In order to specify an accelerator to handle part or all of the computation graph in Android with TFLite, delegates can be used.
In our experiments, we use the XNNPACK, GPU, NNAPI and HEX (Hexagon) delegates.\footnote{\href{https://www.tensorflow.org/lite/performance/delegates}{https://www.tensorflow.org/lite/performance/delegates}}
Each of the considered quantisation schemes performs differently on each processor. For instance, running an FP16 model on the GPU is expected to result in the highest speedup compared to other processors, whereas integer models are best suited for the Hexagon DSP.
Table~\ref{tab:compat} shows the average percentage of the evaluated models' nodes that are delegated through each delegate for the two target devices. \textit{N/S} values mean that execution is not supported by default, whereas zeros mean that the given model could not be executed due to unsupported operations.
Of the available accelerator delegates, only the GPU one has potential to speedup execution in both devices.

\begin{table}[h!]
    \centering
    \vspace{-0.4cm}
    \caption{Delegate Compatibility}
    \label{tab:compat}
    \vspace{-0.2cm}
    \setlength{\tabcolsep}{2pt}
    \resizebox{\columnwidth}{!}{
        \begin{tabular}{ r | c c c c c c | c c c }
            \toprule
            & \multicolumn{6}{c|}{\textbf{Samsung A71}} & \multicolumn{3}{c}{\textbf{Samsung S20 FE}} \\
            \textbf{Variant} & \textbf{CPU} & \multicolumn{2}{c}{\textbf{GPU}} & \multicolumn{2}{c}{\textbf{DSP}} & \textbf{NPU} & \textbf{CPU} & \textbf{GPU} & \textbf{NPU} \\
            & \textbf{XNN} & \textbf{GPU} & \textbf{NNAPI} & \textbf{HEX} & \textbf{NNAPI} & \textbf{NNAPI} & \textbf{XNN} & \textbf{GPU} & \textbf{NNAPI} \\
            \midrule
            \textbf{FP32} & 74.1 & 99.8 & 72.6 & \textit{N/S} & \textit{N/S} & \textit{N/S} & 74.1 & 99.8 & 0   \\
            \textbf{FP16} & 77.6 & 81.8 & 0    & \textit{N/S} & \textit{N/S} & \textit{N/S} & 77.6 & 81.8 & 0   \\
            \textbf{DR8}  & 67.6 & 99.8 & 71.3 & \textit{N/S} & \textit{N/S} & \textit{N/S} & 67.6 & 99.8 & 2.6 \\
            \textbf{FX8}  & 64.0 & 99.8 & 0    & 24.9         & 0            & 1.4          & 64.0 & 99.8 & 1.1 \\
            \bottomrule
        \end{tabular}
    }
    \vspace{-0.4cm}
\end{table}



\vspace{-0.05cm}
\subsection{Benchmarking Details}
For the CPU measurements, we tested the XNNPACK library and multithreading,
while for the GPU and NNAPI delegates, we used 16 bits for computation, when possible.
Prior to taking any measurements, we run the model for a few iterations (1-5) to warm up the processor and decrease measurement variations.
In order to keep the device's temperature as constant as possible and prevent overheating, we run fewer inferences for larger models ($\approx$ 20 runs) than for relatively smaller models ($\approx$ 100 runs).
We also maintained a device idle period of 2-3 minutes between measurements.

\vspace{-0.15cm}
\section{Results}
\label{sec:results}
\vspace{-0.1cm}

\subsection{On-Device Accuracy}
\vspace{-0.05cm}
The reported accuracy in Table~\ref{tab:transformers} comes from evaluation conducted with Python's TFLite Interpreter on an Intel\textsuperscript{\textregistered} Xeon\textsuperscript{\textregistered} CPU.
By evaluating the models on-device, we found that all of the delegate-processor combinations (columns in Table~\ref{tab:compat}) deliver the accuracy presented in Table~\ref{tab:transformers}, \textit{except} for the GPU delegate, where only MobileBERT retains its original accuracy, while for the rest of the discriminative models the accuracy drops drastically (below 35\%).

\vspace{-0.05cm}
\subsection{CPU Performance}
\begin{figure}[h]
    \centering
    \vspace{-0.5cm}
    \includegraphics[width=1.0\columnwidth,trim={0.45cm 3.1cm 1.4cm 2.8cm},clip]{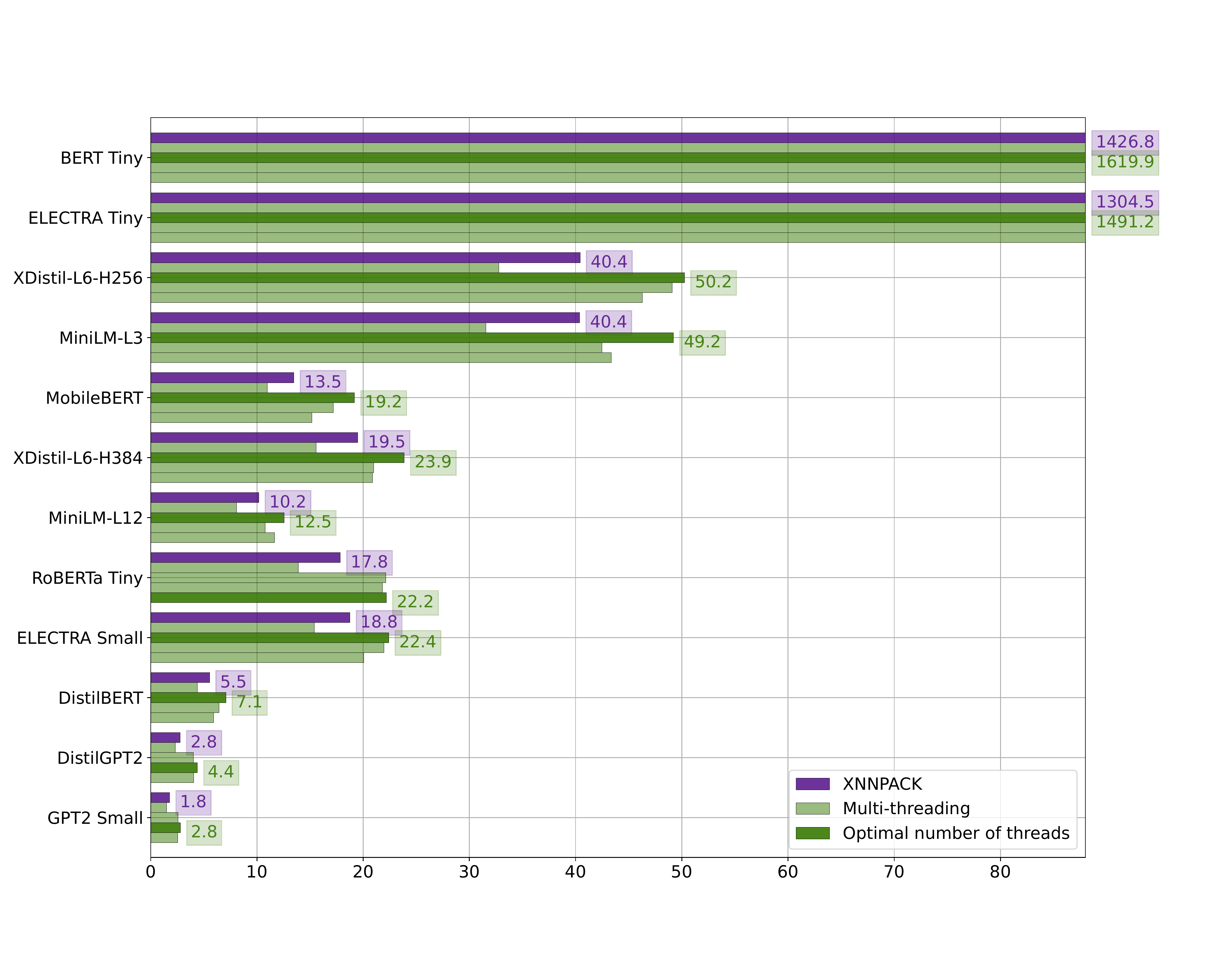}
    \vspace{-0.8cm}
    \caption{CPU throughput (input samples / sec) for Samsung A71.}
    \vspace{-0.45cm}
    \label{fig:cpu_a71}
\end{figure}

\begin{figure}[h]
    \centering
    \vspace{-0.4cm}
    \includegraphics[width=1.0\columnwidth,trim={0.45cm 3.1cm 1.4cm 2.8cm},clip]{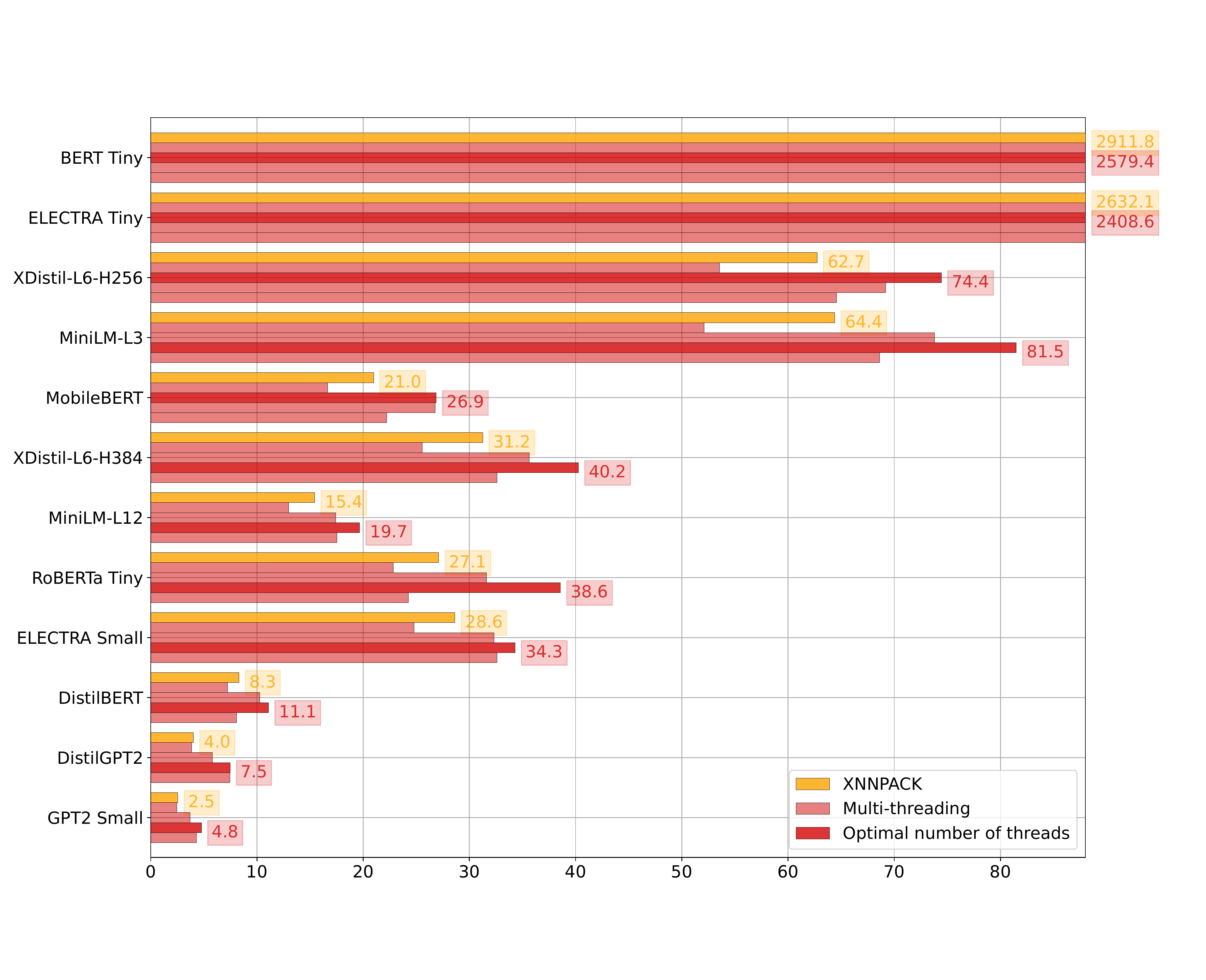}
    \vspace{-0.8cm}
    \caption{CPU throughput (input samples / sec) for Samsung S20 FE.}
    \vspace{-0.1cm}
    \label{fig:cpu_s20}
\end{figure}

\vspace{-0.05cm}
Figures~\ref{fig:cpu_a71} and \ref{fig:cpu_s20} show the impact of the XNNPACK and CPU multithreading on the throughput of all FP32 models for the two devices. Each model's five bars correspond to XNNPACK, 1, 2, 4 and 8 CPU threads when viewed from top to bottom. We observe the following:
\begin{itemize}
    \item Even though XNNPACK is enabled by default in the TFLite Interpreter, it is not the optimal configuration for the majority of models in both devices.
    \item There is no optimal configuration across models or devices. For instance, on S20 FE, the best CPU configuration for ELECTRA Tiny is enabling the XNNPACK delegate, whereas for RoBERTa Tiny, 4 threads will lead to the optimal latency. On A71, instead of enabling the XNNPACK delegate, the best configuration for ELECTRA Tiny is 2 threads.    
    This observation is also valid for the quantised variants of the models.
\end{itemize}

Table~\ref{tab:cpu_quant} shows the effects of quantisation on latency and memory. The memory needed for each FP32 model is also shown in MB for comparison. The values are averaged across the two devices as they were very close. FP16 models achieve the same latency as FP32 models and require at least 70\% \textit{more} memory. In contrast, integer models accelerate the execution and at the same time reduce the memory requirements for the majority of the architectures.

\begin{table}[H]
    \vspace{-0.2cm}
    \centering
    \caption{\small Impact of Quantisation on CPU}
    \vspace{-0.2cm}
    \setlength{\tabcolsep}{4pt}
    \resizebox{\columnwidth}{!}{
        \small
        \begin{tabu}{l | c c c | r c c c}
            \toprule
            \multirow{2}{*}{\textbf{Model}} & \multicolumn{3}{c|}{\textbf{Latency Speedup}} & \multicolumn{1}{c}{\textbf{RAM}} & \multicolumn{3}{c}{\textbf{RAM Reduction}} \\
            & \textbf{FP16} & \textbf{DR8} & \textbf{FX8} & \multicolumn{1}{c}{\textbf{FP32}} & \textbf{FP16} & \textbf{DR8} & \textbf{FX8} \\
            \midrule
            BERT Tiny       & 0.97$\times$ & 0.97$\times$ & 0.42$\times$ & 4.7   & 0.41$\times$ & 0.89$\times$ & 1.28$\times$  \\
            ELECTRA Tiny    & 0.99$\times$ & 0.99$\times$ & 0.39$\times$ & 5.1   & 0.18$\times$ & 0.91$\times$ & 1.13$\times$  \\
            XDistil-L6-H256 & 0.99$\times$ & 1.62$\times$ & 1.51$\times$ & 26.2  & 0.32$\times$ & 2.54$\times$ & 3.03$\times$  \\
            MiniLM-L3       & 1.05$\times$ & 1.86$\times$ & 1.64$\times$ & 30.4  & 0.28$\times$ & 2.84$\times$ & 2.54$\times$  \\
            MobileBERT      & 1.03$\times$ & 1.72$\times$ & -            & 89.4  & 0.59$\times$ & 2.93$\times$ & -             \\
            XDistil-L6-H384 & 0.99$\times$ & 1.82$\times$ & 1.75$\times$ & 52.9  & 0.37$\times$ & 3.07$\times$ & 3.27$\times$  \\
            MiniLM-L12      & 1.02$\times$ & 1.89$\times$ & 1.77$\times$ & 95.2  & 0.47$\times$ & 3.41$\times$ & 3.47$\times$  \\
            RoBERTa Tiny    & 0.99$\times$ & 1.65$\times$ & 1.81$\times$ & 65.0  & 0.37$\times$ & 3.34$\times$ & 3.94$\times$  \\
            ELECTRA Small   & 1.01$\times$ & 1.53$\times$ & 1.82$\times$ & 44.7  & 0.51$\times$ & 3.01$\times$ & 3.01$\times$  \\
            DistilBERT      & 1.06$\times$ & 2.27$\times$ & 2.32$\times$ & 191.9 & 0.46$\times$ & 3.31$\times$ & 3.51$\times$  \\
            DistilGPT2      & -            & -            & -            & 585.2 & -            & -            & -             \\
            GPT2            & 0.98$\times$ & -            & -            & 673.0 & 0.59$\times$ & -            & -             \\
            \bottomrule
            \end{tabu}
    }
    \label{tab:cpu_quant}
\end{table}

\subsection{Accelerators}
Taking into consideration Table~\ref{tab:compat}, we consider execution on the NPUs purposeless, so we calculate the latency speedups
provided by the GPU and DSP only, for the floating-point and fixed-point models, respectively, compared to the best CPU configuration of each model.
Table~\ref{tab:accelerator_speedup} shows the achieved speedups, as well as memory increases (underlined) for each model, except for GPT2, which was found to be accelerator-incompatible.

\begin{table}[H]
    \vspace{-0.4cm}
    \centering
    \caption{\small Accelerator Latency Speedup and Memory Increase}
    \vspace{-0.2cm}
    \setlength{\tabcolsep}{4pt}
    \resizebox{\columnwidth}{!}{
        \small
        \begin{tabu}{l | c c c | c c}
            \toprule
            \multirow{2}{*}{\textbf{Model}} & \multicolumn{3}{c|}{\textbf{Samsung A71}} & \multicolumn{2}{c}{\textbf{Samsung S20 FE}} \\
            & \multicolumn{2}{c}{\textbf{GPU(FP32/16)}} & \textbf{DSP(FX8)} & \multicolumn{2}{c}{\textbf{GPU(FP32/16)}} \\
            \midrule
            BERT Tiny       & 0.08$\times$ & \underline{7.31$\times$} & 0.44$\times$ & 0.06$\times$ & \underline{13.41$\times$} \\
            ELECTRA Tiny    & 0.09$\times$ & \underline{6.12$\times$} & 0.49$\times$ & 0.06$\times$ & \underline{18.16$\times$} \\
            XDistil-L6-H256 & 0.86$\times$ & \underline{3.52$\times$} & 0.83$\times$ & 0.90$\times$ & \underline{5.65$\times$} \\
            MiniLM-L3       & 0.69$\times$ & \underline{3.49$\times$} & 0.72$\times$ & 1.06$\times$ & \underline{4.57$\times$} \\
            MobileBERT      & 0.97$\times$ & \underline{3.52$\times$} & -            & 1.06$\times$ & \underline{4.46$\times$} \\
            XDistil-L6-H384 & 1.24$\times$ & \underline{3.34$\times$} & 0.86$\times$ & 1.20$\times$ & \underline{4.09$\times$} \\
            MiniLM-L12      & 1.22$\times$ & \underline{3.33$\times$} & 0.92$\times$ & 1.29$\times$ & \underline{4.12$\times$} \\
            RoBERTa Tiny    & 1.19$\times$ & \underline{2.79$\times$} & 0.87$\times$ & 1.91$\times$ & \underline{3.43$\times$} \\
            ELECTRA Small   & 1.37$\times$ & \underline{3.63$\times$} & 0.85$\times$ & 1.56$\times$ & \underline{4.69$\times$} \\
            DistilBERT      & 1.22$\times$ & \underline{2.80$\times$} & 0.71$\times$ & 2.77$\times$ & \underline{3.10$\times$} \\
            \bottomrule
            \end{tabu}
    }
    \label{tab:accelerator_speedup}
    \vspace{-0.25cm}
\end{table}

As expected, the DSP is not able to provide any acceleration due to its small compatibility ratio. 
The use of quantisation does not affect the performance of GPU, since all model variants are executed using {\small\texttt{fp16}} arithmetic. Hence, the GPU can only accelerate floating-point models, with the CPU providing higher speedup for integer models (compare with Table~\ref{tab:cpu_quant}).

\vspace{-0.1cm}
\section{Discussion and Future Work}
\label{sec:discussion}
\vspace{-0.05cm}
In the previous section, several conclusions were drawn which provide insights into how execution can be optimised for higher performance. These optimisations can be broadly categorised into two levels: system-level and model-level.

At the system-level, improvements are needed due to three main reasons: (a)~the absence of a global optimal configuration, which stems from the heterogeneity of models and devices, (b)~the dynamic environment of MDs, \textit{e.g.} available resources at runtime, other processes' competing demands, dynamically changing factors (temperature, battery), and (c)~the diverse application performance needs, in terms of accuracy, latency, memory or energy consumption.
Future advancements may include software solutions which are device-dependent and take both model characteristics and performance objectives into account~\cite{contention2022tecs}. Additionally, due to the dynamic environment of MDs, it is important for the system to have access to as many different processors as possible. This enables switching between processors in cases when significant changes in resource availability are observed~\cite{oodin2021smartcomp}.

In our analysis, we found that Transformers are not accelerator-compatible; the GPU can offer marginal speedups but damages accuracy and the DSP and NPUs are mostly not Transformer-compatible.
Influenced by the MobileBERT model, an interesting direction is to consider model-level substitutions, \textit{i.e.} replace complex layers / operations with simpler ones, which have been proved to be more mobile-friendly, such as the replacement of (a)~the GELU activation function (\textit{e.g.} with ReLU), and (b)~Layer Normalisation layers (\textit{e.g.} with elementwise linear transformations or Batch Normalisation).



\vspace{-0.1cm}
\section{Conclusion}
\label{sec:conclusion}
\vspace{-0.1cm}
Our benchmarking results show that despite the increasing adoption of dedicated hardware in modern devices, general-purpose processors, like the CPU, remain highly utilised due to their flexibility in supporting new workloads. 
Our findings
highlight the importance of (a)~research into optimising the Transformer architecture, and (b)~developing accelerators that are more Transformer-friendly. We also stress the need for device- and model-dependent system-level optimisations, since default configurations (\textit{e.g.} XNNPACK) are almost never the optimal choice. We hope our observations and insights can assist future research and development efforts in the area.

\vspace{-0.1cm}
\section{Acknowledgements}
\label{sec:acknowledegments}
\vspace{-0.1cm}
\pichskip{4pt}
\parpic[l]{\includegraphics[width=0.23\columnwidth]{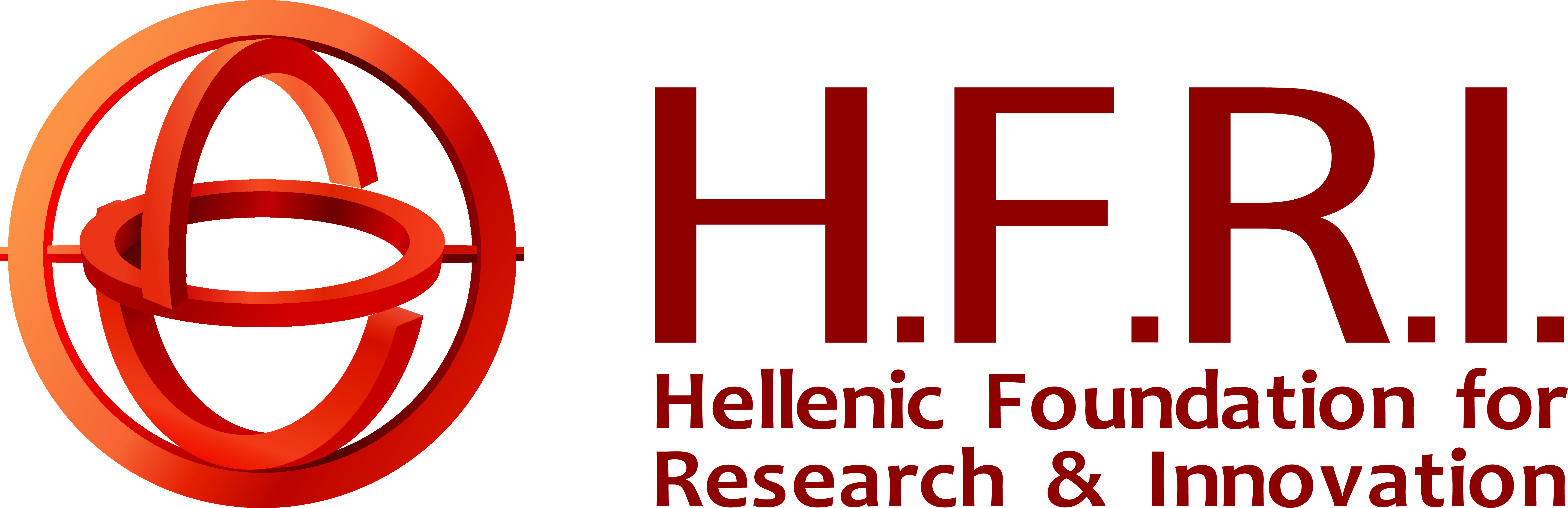}}
\picskip{3}
\noindent\footnotesize{This research work was supported by the Hellenic Foundation for Research and Innovation (HFRI) under the 3rd Call for HFRI PhD Fellowships (Fellowship Number: 5578).}

\bibliographystyle{IEEEtran}
{\footnotesize
\bibliography{references}
}

\end{document}